\documentclass{article}

\usepackage{marvosym} 

\usepackage{iclr2027_conference,times}

\usepackage{amsmath,amsfonts,bm}

\def\eqref#1{equation~\ref{#1}}

\def\1{\bm{1}}

\def\eps{{\epsilon}}

\def\vtheta{{\bm{\theta}}}

\def\mP{{\bm{P}}}

\DeclareMathAlphabet{\mathsfit}{\encodingdefault}{\sfdefault}{m}{sl}
\SetMathAlphabet{\mathsfit}{bold}{\encodingdefault}{\sfdefault}{bx}{n}

\def\sA{{\mathbb{A}}}

\def\sG{{\mathbb{G}}}

\def\sS{{\mathbb{S}}}

\DeclareMathOperator*{\argmax}{arg\,max}

\usepackage{hyperref}
\usepackage{url}
\usepackage{booktabs}
\usepackage{multirow}
\usepackage{graphicx}
\usepackage{amsmath,amssymb,mathtools}
\usepackage{microtype}
\usepackage{xcolor}
\usepackage{enumitem}
\usepackage{adjustbox}
\usepackage{array}
\usepackage{longtable}
\usepackage{caption}
\usepackage{subcaption}
\usepackage{float}
\usepackage{placeins}
\usepackage{wrapfig}
\usepackage{algorithm}
\usepackage{algpseudocode}
\usepackage[T1]{fontenc}
\usepackage[utf8]{inputenc}
\hypersetup{colorlinks=true,citecolor=blue,linkcolor=blue,urlcolor=blue}
\setlist{nosep,leftmargin=1.4em}
\newcommand{\papertableformat}{%
  \footnotesize
  \setlength{\tabcolsep}{4pt}%
  \renewcommand{\arraystretch}{1.06}%
}

\newcommand{\StaticIV}{\textsc{Static-IV}}
\newcommand{\DIRR}{\textsc{DIR-R}}

\newcommand{\Gset}{\sG}                 
\newcommand{\Sset}{\sS}                 
\newcommand{\Aset}{\sA}                 


\title{Storage Is Not Strategy: State-Conditioned Support Control for LLM Unlearning}

\iclrfinalcopy

\author{
Tianhao Qian$^{1}$,
Ziming Hong$^{2}$,
Chongyang Gao$^{3,4}$,
Kezhen Chen$^{4}$,
Lixu Wang$^{1,*}$
\\[5pt]
$^{1}$ The Chinese University of Hong Kong, Shenzhen \\
$^{2}$ The University of Sydney \\
$^{3}$ Northwestern University \\
$^{4}$ Analogy AI, Inc. \\[2pt]
$^{*}$Corresponding authors
}

\date{}

\fancypagestyle{preprintpages}{%
    \fancyhf{}%
    \fancyhead[L]{\textit{Preprint}}%
    \fancyfoot[C]{\thepage}%
}
\fancypagestyle{firstpage}{%
    \fancyhf{}%
    \fancyfoot[C]{\thepage}%
}

\begin{document}

\noindent
\includegraphics[width=0.35\textwidth]{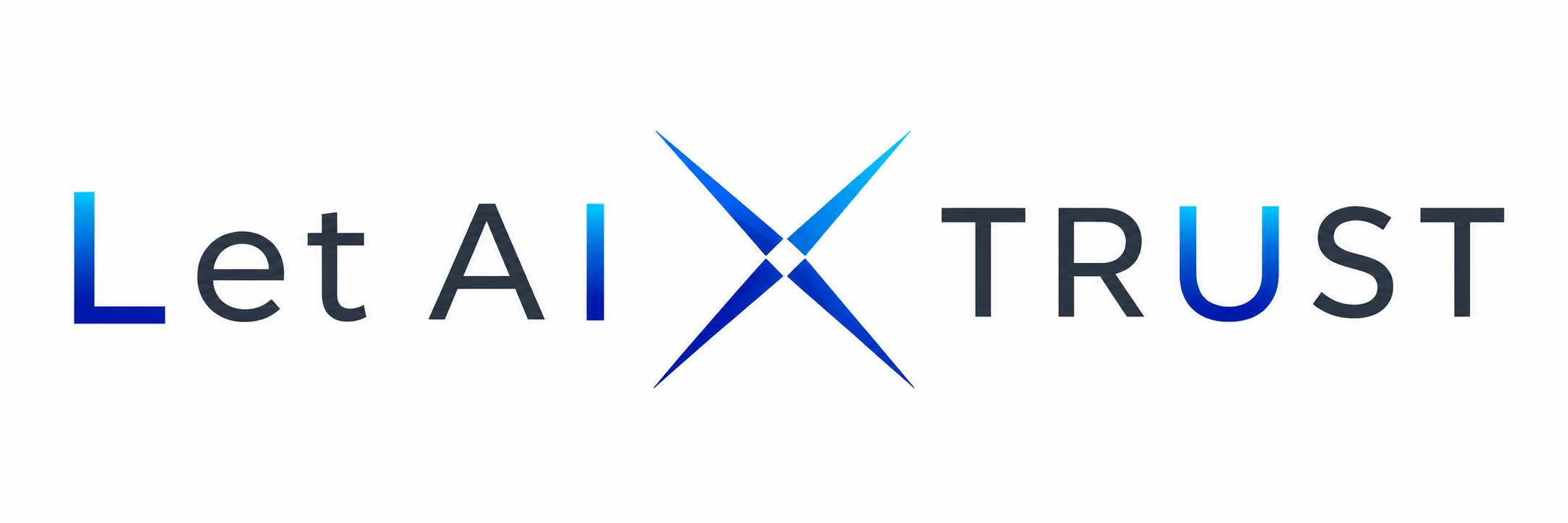}

\vspace{-25pt}
\noindent\rule{\linewidth}{0.4pt}
\vspace{0.6em}

\maketitle
\pagestyle{preprintpages}
\thispagestyle{firstpage}

\begin{abstract}
Many localized large language model (LLM) unlearning methods select a small parameter subset from a localization signal and keep it fixed during optimization. The parameters most associated with a target, however, need not be the best ones to update, and candidate interventions can change value as optimization proceeds. In a controlled experiment, a storage-localization score reaches an area under the receiver operating characteristic curve (AUROC) of $0.981$, yet storage identity agrees with the better intervention on only $17/36$ targets, while low-rank adaptation (LoRA) wins $35/36$. We introduce \emph{Intervention Score}, which ranks editable groups by the predicted effect of the actual unlearning update while accounting for collateral damage, and use it to form the static intervention-value baseline (\StaticIV{}). We then introduce selective dynamic intervention re-ranking (\DIRR{}), which revisits that subset only when a calibrated probe justifies the comparison. On the Natural-TOFU dataset, our method has positive descriptive margins in $19/20$ comparisons between methods and objectives, although several are near zero. On the LACUNA localization-precision benchmark, our mean terminal utility is higher in all six negative preference optimization (NPO) and SimNPO comparisons: NPO margins range from $+0.431$ to $+0.848$, and SimNPO margins range from $+0.503$ to $+0.571$. The gradient-difference (GradDiff) objective reveals substantial field dependence. Relative to \StaticIV{}, the primary four-field GradDiff evaluation has six wins, six ties, and no losses, with mean and median paired gains of $+0.165$ and $+0.0025$. The evidence supports separating localization, initial intervention selection, and checkpoint-dependent support revision.
\end{abstract}

\section{Introduction}

Machine unlearning studies how to remove the effect of specified training data from a trained model while preserving utility on data that should be retained \citep{cao2015unlearning,bourtoule2021machine,zhao2025exploring}. A natural reference is retraining without the data to be forgotten, but retraining is prohibitively expensive for LLMs \citep{maini2024tofu,shi2024muse}. Practical LLM unlearning therefore modifies an existing model in place. Recent work develops direct gradient objectives and preference losses, including negative preference optimization (NPO) and its reference-free simplification SimNPO \citep{jang2023knowledge,zhang2024npo,fan2024simnpo}. A complementary line asks where the update should act. Localized LLM unlearning selects a small parameter subset from localization or attribution signals and restricts optimization to that subset \citep{jia2024wagle,tian2024memflex,guo2025mechanistic,lee2025localization}.

Localization-first designs often rely on two operational assumptions. The first treats localization-derived importance as a proxy for where a particular unlearning objective should intervene, although localization quality does not necessarily predict intervention efficacy \citep{hase2023localization,lee2025localization}. The second keeps the selected parameter subset fixed as optimization proceeds, even though continual-unlearning work adapts masks dynamically and post-training analysis warns that static mechanistic localization can become stale as parameters evolve \citep{wuerkaixi2025adaptive,chen2026oldmaps}. We therefore separate the initial intervention subset from the decision to revise it after optimization changes the model.

Our controlled experiment directly tests the first assumption by separating localization accuracy from intervention effectiveness. We construct synthetic facts for which the parameter interface used to inject each fact is known by design. A storage-localization score identifies this known interface with AUROC $0.981$ and residualized AUROC $0.926$, indicating accurate localization. Yet it predicts the better intervention on only $17$ of $36$ targets, while an alternative LoRA interface \citep{hu2022lora} achieves the better outcome on $35$ of $36$. Accurate localization does not by itself identify where a particular unlearning objective should intervene. This choice is also objective-dependent. On our Natural-TOFU setting, derived from TOFU \citep{maini2024tofu}, equal-size subsets selected independently under NPO and SimNPO \citep{zhang2024npo,fan2024simnpo} have a Jaccard overlap of about $0.143$. Both observations motivate choosing the initial subset according to the expected effect of the actual update instead of localization alone. Running every candidate intervention to completion would be prohibitively expensive, so we introduce \emph{Intervention Score} as an efficient proxy. The score estimates how the update induced by the chosen objective would act on each group. It rewards predicted forgetting, penalizes predicted damage to retained and protected behavior, and discounts nonspecific model changes. Ranking under a fixed parameter budget yields an objective-conditioned initial subset. Keeping it fixed defines \StaticIV{} and isolates initial selection quality. Because even a strong initial subset can become less suitable as the model changes, selective \DIRR{} revisits it at a later checkpoint and runs the complete continuation comparison only when a calibrated probe indicates potential value.

We evaluate the two decisions in complementary settings. On the Natural-TOFU dataset, Intervention Score with \DIRR{} is compared with ten parameter-selection baselines under NPO and SimNPO; its descriptive mean margins are positive in $19/20$ comparisons, with several close to zero. The LACUNA localization-precision benchmark \citep{boglioni2026lacuna} provides direct comparisons with selected baselines. After selecting three objective-specific baselines from a ten-method panel, our method has higher mean terminal utility in all six NPO and SimNPO comparisons. NPO margins range from $+0.431$ to $+0.848$, and SimNPO margins range from $+0.503$ to $+0.571$. GradDiff exposes a boundary: static parameter-subset rankings change substantially across data fields, and several static methods remain ahead in the complete four-field primary. In the \DIRR{} ablation, the aggregate mean difference relative to \StaticIV{} is positive in all six evaluation sets; the primary GradDiff evaluation has mean gain $+0.1646$, median paired gain $+0.0025$, and six wins, six ties, and no losses. The results treat initial selection and checkpoint-dependent revision as distinct empirical questions without claiming uniform dominance.

\textbf{Our contributions are:}
\begin{itemize}[leftmargin=*]
\item \textbf{A controlled separation of localization and intervention.} We show that highly accurate storage localization need not identify the better unlearning intervention: despite storage AUROC $0.981$, storage identity agrees with the better intervention on only $17/36$ targets.

\item \textbf{Objective-conditioned initial parameter selection.} We introduce Intervention Score and its fixed-subset realization \StaticIV{}. Against ten parameter-selection methods under NPO and SimNPO on the Natural-TOFU dataset, Intervention Score with \DIRR{} has positive descriptive margins in $19/20$ comparisons.

\item \textbf{Checkpoint-dependent support revision.} We introduce selective \DIRR{}, which conditionally revisits the initial subset rather than recomputing alternatives throughout optimization. The LACUNA benchmark reports positive mean margins in all six NPO and SimNPO baseline comparisons, while the ablation of \DIRR{} against \StaticIV{} has a positive aggregate mean difference in every reported evaluation set.
\end{itemize}

\section{Related Work}
\label{sec:related-work}

\paragraph{Unlearning.}
Classical machine unlearning studies deletion-efficient learning, certified removal, selective forgetting, and retraining-oriented guarantees \citep{ginart2019making,guo2020certified,golatkar2020eternal,sekhari2021remember}. LLM-specific work extends this setting with privacy-oriented dememorization, direct parameter updates, representation-level removal, parameter-efficient methods, causal formulations, and inference-time alternatives \citep{kassem2023dememorization,chen2023efficient,ji2024uld,yao2024pretrained,pawelczyk2024incontext,liu2024causal,cha2025loku,ding2025unified,gao2025continual}. Evaluation work additionally stresses that apparent forgetting can be fragile under stronger metrics, obfuscation tests, or relearning \citep{hu2025obfuscating,wang2025evaluation,fan2025relearning}. Our work fixes the unlearning objective and studies a different decision: which parameters that objective should update, and whether that choice should later change.

\paragraph{Localization.}
Model editing uses direct parameter modification, learned editors, retrieval mechanisms, or localized weight updates to change factual behavior \citep{decao2021editing,mitchell2022mend,mitchell2022serac,meng2022rome,meng2023memit}; later work further questions static fact-level localization assumptions \citep{chen2025query}. Localized unlearning inherits the same tension. WAGLE and MemFlex choose update regions from attribution or gradient evidence \citep{jia2024wagle,tian2024memflex}, while mechanistic and benchmark studies test how localization affects robustness and intervention efficacy \citep{guo2025mechanistic,lee2025localization,boglioni2026lacuna}. We treat localization evidence as an input to an intervention decision rather than as the decision target itself.

\section{Preliminaries and Problem Formulation}
\label{sec:preliminaries}
\label{sec:problem}
Table~\ref{tab:notation} in Appendix~\ref{app:method-details} collects the notation used below.

\subsection{Localized unlearning as support-constrained optimization}
Let $f_{\vtheta_0}$ be the model before unlearning, with forget set $\mathcal D_f$ and retain set $\mathcal D_r$. We fix an unlearning objective $\mathcal L_{\mathrm{unl}}$, its optimizer and minibatch sequence, and a horizon of $T$ optimization steps. Let $\Gset$ be a finite collection of editable parameter groups. For $g\in\Gset$, $\vtheta_g$ denotes the coordinates in that group. A support $\Sset\subseteq\Gset$ specifies which groups may change during unlearning; all coordinates outside $\Sset$ remain frozen. If $\mP_{\Sset}$ projects onto the active coordinates, a support-constrained update has the form
\begin{equation}
\vtheta_{k+1}^{\Sset}
=\vtheta_k^{\Sset}-\eta_k\mP_{\Sset}\mathbf u_k^{\Sset},
\label{eq:problem-support-update}
\end{equation}
where $\mathbf u_k^{\Sset}$ is the optimizer direction along the trajectory induced by support $\Sset$. Thus, the support is an intervention variable: it determines where the declared unlearning algorithm is allowed to act.

The editable-parameter cost of a support is
\begin{equation}
c(\Sset)=\sum_{g\in\Sset}\dim(\vtheta_g),
\label{eq:support-cost}
\end{equation}
and the feasible family under budget $B$ is
\begin{equation}
\mathcal F_B=\{\Sset\subseteq\Gset:c(\Sset)\le B\}.
\label{eq:feasible-family}
\end{equation}
For support-comparison experiments, a terminal model is scored by $J(\vtheta)=G_F(\vtheta)-D_{\mathrm{coll}}(\vtheta)$, where $G_F$ measures forgetting gain and $D_{\mathrm{coll}}\ge0$ measures collateral degradation. Larger $J$ is better. The objective, optimizer, minibatches, budget, and evaluator are held fixed when supports are compared.

\subsection{From localization to intervention selection}
A localization method asks which parameter groups are associated with the information to be forgotten. That is a descriptive question. Choosing a support asks a different question: if the same unlearning procedure is allowed to update only a particular set of groups, which choice leads to the best terminal outcome? A group may therefore receive a strong localization signal without being the best place to intervene under the actual objective, budget, and optimization trajectory.

For a fixed support $\Sset\in\mathcal F_B$, let $\vtheta_T^{\Sset}$ be the endpoint obtained by running the fixed unlearning procedure from $\vtheta_0$ for all $T$ steps while updating only $\Sset$. Its \emph{initial intervention value} is
\begin{equation}
V_0(\Sset)=J(\vtheta_T^{\Sset}),
\label{eq:initial-intervention-value}
\end{equation}
and the ideal initial support is
\begin{equation}
\Sset_0^\star\in\argmax_{\Sset\in\mathcal F_B}V_0(\Sset).
\label{eq:static-target}
\end{equation}
Computing Eq.~\ref{eq:static-target} directly would require full unlearning runs for many feasible supports. The first problem is therefore to select a practical $\Sset_0\in\mathcal F_B$ that is informed by the unlearning objective without exhaustively evaluating terminal outcomes.

\subsection{Checkpoint-conditioned support revision}
The initial support is chosen at $\vtheta_0$, but optimization changes the model and its optimizer state. After following $\Sset_0$ to step $t$, let $\mathcal C_t$ denote a restorable checkpoint. Restoring $\mathcal C_t$ reinstates $\vtheta_t$, the optimizer and scheduler state, the position in the minibatch sequence, and the random-number-generator state. These quantities are restored identically for every continuation, so the editable support is the only intervention that changes.

For a continuation support $\Sset\in\mathcal F_B$, let $\vtheta_T(\Sset;\mathcal C_t)$ be the endpoint obtained by restoring $\mathcal C_t$ and running the remaining steps with updates restricted to $\Sset$. Its \emph{continuation value} is
\begin{equation}
V_t(\Sset\mid\mathcal C_t)
=J\!\left(\vtheta_T(\Sset;\mathcal C_t)\right),
\label{eq:continuation-value}
\end{equation}
with ideal checkpoint-dependent continuation
\begin{equation}
\Sset_t^\star(\mathcal C_t)
\in\argmax_{\Sset\in\mathcal F_B}V_t(\Sset\mid\mathcal C_t).
\label{eq:continuation-target}
\end{equation}
There is no requirement that $\Sset_t^\star(\mathcal C_t)$ coincide with the support preferred at initialization. The second problem is therefore computational: after observing $\mathcal C_t$, determine whether the current support is still adequate or whether a comparison of alternative continuations is worth its cost. Our method addresses these two decisions: initial intervention selection and checkpoint-conditioned revision.

\section{Methodology}
\label{sec:method}
\subsection{Method overview}
The method separates support selection from support revision. At initialization, \emph{Intervention Score} ranks parameter groups by the predicted effect of the actual unlearning update and produces support $\Sset_0$. \StaticIV{} uses this support for the entire trajectory. Selective \DIRR{} adds a decision at checkpoint $\mathcal C_t$: a short comparison determines whether the frozen continuation candidates should be evaluated to completion. Algorithm~\ref{alg:dirr} gives the full procedure.

This construction is narrower than solving Eqs.~\ref{eq:static-target} and~\ref{eq:continuation-target} globally. Intervention Score is a local surrogate for terminal intervention value, and \DIRR{} searches a small, prespecified family around the initial support. The method makes both decisions testable under a fixed training procedure; it does not claim global support optimality.

\subsection{Intervention Score: objective-conditioned initial parameter selection}
\label{sec:static-support-selection}
\label{sec:intervention-score}
A localization score is typically derived from how the forget data are represented or attributed in the current model. Intervention Score instead asks how a candidate group would participate in the update induced by $\mathcal L_{\mathrm{unl}}$. Let $\mathbf u_0$ be the preconditioner-free direction induced by the unlearning objective at $\vtheta_0$, defined so that a small objective step is $\vtheta_0-\eta\mathbf u_0$. Let $\mP_g$ project onto group $g$.

We use four outcome-blind diagnostic roles $q\in\{F,R,P,N\}$: forget, retain, protected behavior, and neutral behavior. The forget diagnostic measures whether the restricted update moves in a direction favorable to forgetting. The retain and protected diagnostics flag predicted collateral degradation, with the protected role covering behavior that is not represented by the ordinary retain set. The neutral diagnostic measures whether the same update produces broad, nonspecific movement. These diagnostics are used only to construct the parameter-selection rule and do not read the terminal outcomes used for evaluation. Let $\ell_q$ be a diagnostic loss and $\mathbf g_q=\nabla_{\vtheta}\ell_q(\vtheta_0)$.

The score follows from the first-order change produced by a restricted step. For small $\eta$,
\begin{equation}
\ell_q(\vtheta_0-\eta\mP_g\mathbf u_0)
=\ell_q(\vtheta_0)-\eta\left\langle\mP_g\mathbf g_q,\mP_g\mathbf u_0\right\rangle+O(\eta^2).
\label{eq:intervention-taylor}
\end{equation}
We therefore define the signed first-order effect
\begin{equation}
e_{g,q}=-\left\langle \mP_g\mathbf g_q,\mP_g\mathbf u_0\right\rangle.
\label{eq:group-alignment}
\end{equation}
The diagnostics are oriented so that $e_{g,F}>0$ predicts useful forgetting, whereas $e_{g,R}>0$ and $e_{g,P}>0$ predict collateral degradation; $|e_{g,N}|$ measures nonspecific movement. Intervention Score is
\begin{equation}
s_{\mathrm{int}}(g)=
\frac{e_{g,F}-\max\{e_{g,R},e_{g,P},0\}}
{|e_{g,N}|+\eps},
\label{eq:intervention-score}
\end{equation}
where $\eps>0$ is a fixed numerical stabilizer. The numerator rewards predicted forgetting while subtracting the larger predicted collateral effect. The denominator downweights groups whose predicted effect is broad rather than forget-directed. This score is used as a ranking surrogate for $V_0$; it is not assumed to equal the terminal value of a support.

Groups are ranked by $s_{\mathrm{int}}$ and added while respecting $c(\Sset)\le B$, producing $\Sset_0$. In Natural-TOFU and LACUNA, an atomic group is one input column of an MLP down-projection matrix. Groups within a model therefore have equal parameter count, so ranking and packing reduce to taking the highest-scoring groups up to the budget. The candidate geometry and editable-parameter budget are the same for all parameter-selection methods in a comparison.

\subsection{\StaticIV{}: fixing the initial support}
\label{sec:static-iv}
\StaticIV{} keeps $\Sset_0$ active for all $T$ steps. At step $k$, the objective and optimizer produce the support-conditioned direction $\mathbf u_k^{\Sset_0}$ and step size $\eta_k$, giving
\begin{equation}
\vtheta_{k+1}^{\Sset_0}=\vtheta_k^{\Sset_0}-\eta_k\mP_{\Sset_0}\mathbf u_k^{\Sset_0}.
\label{eq:static-iv-update}
\end{equation}
This variant answers a specific ablation question: what is obtained from the objective-conditioned initial parameter selection if the support is never reconsidered? It also supplies the current \StaticIV{} continuation against which \DIRR{} evaluates possible revisions.

\subsection{Selective \DIRR{}: revising the support when useful}
\label{sec:dirr-section}
\label{sec:dirr}
The score that produced $\Sset_0$ was computed at initialization. By step $t$, the parameters, optimizer state, and remaining horizon have changed, so the initial ranking need not remain best for the continuation problem in Eq.~\ref{eq:continuation-target}. Resolving that problem at every checkpoint would defeat the purpose of localized unlearning. \DIRR{} therefore separates whether to run a continuation comparison from which support to choose after a trigger.

\paragraph{Local candidate family.}
Order the groups in $\Sset_0$ from lowest to highest Intervention Score as $g^-_1,\ldots,g^-_M$, and unselected groups from highest to lowest as $g^+_1,g^+_2,\ldots$. For replacement fraction $\rho$, let $k(\rho)=\lfloor\rho M\rfloor$ and define
\begin{equation}
\Sset^{(\rho)}=
\left(\Sset_0\setminus\{g^-_1,\ldots,g^-_{k(\rho)}\}\right)
\cup\{g^+_1,\ldots,g^+_{k(\rho)}\}.
\label{eq:candidate-support}
\end{equation}
We freeze $\Aset=\{0,0.01,0.025,0.05,0.10\}$, where $\rho=0$ is the unchanged support. In the primary equal-cost geometry, every exchange preserves the editable-parameter count exactly. With heterogeneous group costs, the implementation must instead enforce $c(\Sset^{(\rho)})=c(\Sset_0)$ explicitly.

\paragraph{Short probe and gate.}
Running every candidate to step $T$ is expensive, so the gate first compares the current support with one fixed probe action $\rho_p\in\Aset\setminus\{0\}$. For support $\Sset$, let $\vtheta_{t+d}(\Sset;\mathcal C_t)$ denote the state reached after exactly $d$ additional steps from the same checkpoint, using the same minibatches and randomness. Define
\begin{equation}
\widetilde V_t^{(d)}(\rho\mid\mathcal C_t)
=J\!\left(\vtheta_{t+d}(\Sset^{(\rho)};\mathcal C_t)\right),
\label{eq:probe-value}
\end{equation}
and the nonnegative probe signal
\begin{equation}
s_t=\max\{0,\widetilde V_t^{(d)}(\rho_p\mid\mathcal C_t)-\widetilde V_t^{(d)}(0\mid\mathcal C_t)\}.
\label{eq:selective-signal}
\end{equation}
The probe does not choose the final support. It only asks whether the observed checkpoint provides enough evidence to justify a more expensive comparison. If $s_t\le\tau^\star$, the run continues with $\Sset^{(0)}$ and no full continuation comparison is performed.

\paragraph{Exhaustive continuation comparison after a trigger.}
If $s_t>\tau^\star$, every candidate is restored from the same $\mathcal C_t$ and run to $T$ under the same future minibatches and randomness. \DIRR{} chooses
\begin{equation}
\widehat\rho_t
=\argmax_{\rho\in\Aset}V_t(\Sset^{(\rho)}\mid\mathcal C_t),
\label{eq:exhaustive-action-rule}
\end{equation}
with ties broken toward smaller $\rho$. Including $\rho=0$ means the exhaustive comparison can retain the \StaticIV{} continuation whenever every proposed exchange is worse. This guarantee is only relative to the frozen candidate family; it is not a claim of global optimality over $\mathcal F_B$.

\paragraph{Threshold calibration and compute reference.}
The threshold $\tau^\star$ is selected using development examples only. At each development checkpoint, the probe signal is paired with the positive available gain inside the frozen candidate family. Among thresholds in a prespecified trigger-rate range, calibration favors the threshold that captures the most positive available gain per unit of additional step-equivalent compute. Final outcomes do not alter $\tau^\star$, $\rho_p$, $d$, $\Aset$, the budget, or the unlearning objective. Natural-TOFU uses $T=200$, $t=160$, $d=20$, and $\rho_p=0.01$; the calibration equations are given in Appendix~\ref{app:method-details}.

For context, the $T=200$ exhaustive multi-checkpoint reference evaluates all five actions from checkpoints $\{0,40,80,120,160\}$. It requires $5(200+160+120+80+40)=3000$ continuation steps in addition to the committed $200$-step trajectory, or $16\times$ normalized step-equivalent compute. This exhaustive construction is used only as a compute-heavy reference and is distinct from the single-checkpoint exhaustive comparison invoked by deployed \DIRR{}.

\begin{algorithm}[t]
\caption{Objective-conditioned support selection with selective \DIRR{}}
\label{alg:dirr}
\scriptsize
\begin{algorithmic}[1]
\Require initial model $\vtheta_0$, forget and retain data, objective $\mathcal L_{\mathrm{unl}}$, diagnostics $\{\ell_q\}_{q\in\{F,R,P,N\}}$, groups $\Gset$, budget $B$, horizon $T$
\Require checkpoint $t$, probe length $d$, actions $\Aset$, probe action $\rho_p$, calibrated threshold $\tau^\star$
\State Compute $\mathbf u_0$ and $\mathbf g_q=\nabla_{\vtheta}\ell_q(\vtheta_0)$
\For{each $g\in\Gset$}
  \State $e_{g,q}\gets-\langle \mP_g\mathbf g_q,\mP_g\mathbf u_0\rangle$ for $q\in\{F,R,P,N\}$
  \State $s_{\mathrm{int}}(g)\gets\bigl(e_{g,F}-\max\{e_{g,R},e_{g,P},0\}\bigr)/( |e_{g,N}|+\eps)$
\EndFor
\State Select $\Sset_0$ under $B$; construct $\{\Sset^{(\rho)}:\rho\in\Aset\}$ with $\Sset^{(0)}=\Sset_0$
\State Run with $\Sset_0$ to $t$; save $\mathcal C_t$; probe $\rho=0$ and $\rho_p$ from independent restorations
\State $s_t\gets\max\{0,\widetilde V_t^{(d)}(\rho_p\mid\mathcal C_t)-\widetilde V_t^{(d)}(0\mid\mathcal C_t)\}$
\If{$s_t\le\tau^\star$} \State \Return $\vtheta_T(\Sset^{(0)};\mathcal C_t)$
\Else
  \State Evaluate every $\rho\in\Aset$ from independent restorations; set $\widehat\rho_t\gets\argmax_{\rho\in\Aset}V_t(\Sset^{(\rho)}\mid\mathcal C_t)$
  \State \Return $\vtheta_T(\Sset^{(\widehat\rho_t)};\mathcal C_t)$
\EndIf
\end{algorithmic}
\end{algorithm}

\section{Experiments}
\label{sec:experiments}

\paragraph{Overview.}
We test localization versus intervention in a controlled setting, breadth on Natural-TOFU, comparisons and heterogeneity on LACUNA, and checkpoint-dependent revision in the final ablation.

\paragraph{Metrics.}
The terminal utility is $J=G_F-D_{\mathrm{coll}}$, where larger values mean more forgetting gain after benchmark-specific nonnegative collateral damage. For paired runs, $\Delta J(A,B)=n^{-1}\sum_i[J_i(A)-J_i(B)]$, so positive values favor $A$. We also report paired medians, wins/ties/losses (W/T/L), and leave-one-out (LOO) ranges for GradDiff. \DIRR{} metrics are defined in Appendix~\ref{app:metric-definitions}.

\subsection{Experimental Protocol}
\label{sec:experimental-protocol}

\paragraph{Setup.}
Our Natural-TOFU dataset, derived from TOFU \citep{maini2024tofu}, uses Llama 3 8B \citep{grattafiori2024llama3} and Gemma 2 2B \citep{gemmateam2024gemma2} under NPO and SimNPO \citep{zhang2024npo,fan2024simnpo}, with approximately six million editable multilayer-perceptron (MLP) down-projection scalars. The LACUNA localization-precision benchmark \citep{boglioni2026lacuna} uses frozen OLMo 3 7B \citep{olmo2025olmo3} on Email Address, Driver's License, Birth City, and Phone Number under NPO, SimNPO, and gradient difference (GradDiff) \citep{yao2024pretrained}. We compare weight-attribution-guided LLM unlearning (WAGLE) \citep{jia2024wagle}, MemFlex \citep{tian2024memflex}, Fisher-initialized low-rank adaptation (FILA) \citep{cha2025loku}, activation-patching MLP (AP-MLP) and Activation-Down \citep{lee2025localization}, gradient-based adaptive unlearning (GRAIL) \citep{kim2025grail}, and Projected Causal \citep{liu2024causal,guo2025mechanistic}. Random, Activation, and Storage are outcome-blind controls defined in this study. All methods use the same editable geometry when available.

\paragraph{Protocol.}
Natural-TOFU combines a completed parameter-selection panel with later DIR-R evaluation runs, so its reported margins are descriptive summaries rather than direct paired estimates. For LACUNA, baseline identities are fixed before outcome evaluation and compared methods use the same field, objective, seed, parameter budget, and evaluator. GradDiff's four-field aggregate is primary; the three-field exclusion of Birth City is a sensitivity analysis defined after observing the field effect. Missing runs are not imputed, and DIR-R settings are frozen before final outcomes are read; Appendix~\ref{app:complete-results} gives the audit.

\subsection{Controlled Separation Between Localization And Intervention Selection}
\label{sec:controlled-storage-action}

\paragraph{Localization.}
Synthetic facts are injected through known parameter interfaces in Qwen2.5-7B \citep{qwen2025qwen25}. Without using intervention outcomes, the locator reaches AUROC $0.98148$, residualized AUROC $0.92593$, and the correct direction on $17/18$ blind pairs.

\begin{table}[H]
\vspace{-0.55\baselineskip}
\centering
\caption{Controlled separation between localization and intervention.}
\label{tab:controlled-main}
\footnotesize
\setlength{\tabcolsep}{2.8pt}
\renewcommand{\arraystretch}{0.96}
\begin{tabular}{@{}ll@{}}
\toprule
Quantity & Result \\
\midrule
Storage AUROC & $0.98148$ \\
Residualized storage AUROC & $0.92593$ \\
Correct within-pair storage direction & $17/18$ \\
LoRA intervention wins & $35/36$ \\
Storage/best-intervention agreement & $17/36$ ($47.2\%$) \\
LoRA$-$backbone paired difference & $+0.1905$ $[0.1570,0.2242]$ \\
Storage-score / paired-difference corr. & $-0.2302$ \\
\bottomrule
\end{tabular}
\vspace{-0.35\baselineskip}
\end{table}

\paragraph{Intervention.}
Across 36 NPO targets, LoRA wins $35/36$, while storage identity agrees with the better intervention on only $17/36$ ($47.2\%$). The mean effect is $+0.1905$ ($95\%$ interval $[0.1570,0.2242]$), and localization score correlates negatively with the LoRA-minus-backbone paired difference ($-0.2302$). \textbf{\textit{Accurate localization does not reliably identify the better intervention in this controlled setting.}}
\par\vspace{0.15\baselineskip}

\FloatBarrier
\subsection{Natural-TOFU: Broad Method Comparison}
\label{sec:natural-main}

\begin{wraptable}[10]{r}{0.50\textwidth}
\vspace{-0.65\baselineskip}
\centering
\caption{Natural-TOFU margins ($\Delta J$); positive values favor ours.}
\label{tab:natural-main}
\scriptsize
\setlength{\tabcolsep}{1.5pt}
\renewcommand{\arraystretch}{0.90}
\begin{tabular}{@{}lcc@{}}
\toprule
Baseline & NPO $\Delta J$ & SimNPO $\Delta J$ \\
\midrule
Random & $+0.004799$ & $+0.006974$ \\
Activation & $+0.041006$ & $+0.154743$ \\
WAGLE & $+0.136450$ & $+0.297768$ \\
Storage & $+0.032802$ & $+0.162268$ \\
MemFlex & $+0.094780$ & $+0.210191$ \\
FILA-relative Fisher & $+0.003390$ & $-0.000799$ \\
AP-MLP & $+0.029772$ & $+0.098141$ \\
GRAIL & $+0.002172$ & $+0.000719$ \\
Activation-Down & $+0.039595$ & $+0.149278$ \\
Projected Causal & $+0.110948$ & $+0.131786$ \\
\bottomrule
\end{tabular}
\vspace{-0.35\baselineskip}
\end{wraptable}

\paragraph{Scope.}
The Natural-TOFU dataset compares Intervention Score with \DIRR{} against ten parameter-selection methods under NPO and SimNPO. Because the parameter-selection and \DIRR{} components come from different final evaluation sets, Table~\ref{tab:natural-main} is descriptive rather than a paired state-of-the-art estimate.

\paragraph{Results.}
\textbf{\textit{Margins are positive in $19/20$ comparisons, but several are close to zero.}} The largest is $+0.297768$ against WAGLE under SimNPO; the only negative is the $-0.000799$ near-tie with FILA-relative Fisher.

\FloatBarrier
\subsection{LACUNA: Comparisons And GradDiff Heterogeneity}
\label{sec:lacuna-main}

\paragraph{Comparisons.}
Table~\ref{tab:lacuna-main} reports the six NPO/SimNPO comparisons and the GradDiff sensitivity.

\begin{table}[!t]
\centering
\caption{LACUNA results. Panel A gives NPO/SimNPO comparisons. In Panel B, DIR-R is the reference with mean $J=106.591877$; all remaining columns report paired DIR-R-minus-baseline statistics, so no self-comparison row is shown for DIR-R. The four-field primary is in Appendix~\ref{app:graddiff}.}
\label{tab:lacuna-main}
\footnotesize
\renewcommand{\arraystretch}{0.98}

\begin{adjustbox}{max width=\textwidth}
\begin{tabular}{@{}lcccl@{}}
\toprule
\multicolumn{5}{@{}l}{Panel A: NPO/SimNPO comparison} \\
Objective / baseline & Ours & Baseline & $\Delta J$ & Wins/Losses \\
\midrule
NPO / Projected Causal & 6.193908 & 5.763156 & $+0.430752$ & 8/4 \\
NPO / MemFlex & 6.193908 & 5.502749 & $+0.691159$ & 10/2 \\
NPO / Activation exact & 6.193908 & 5.346050 & $+0.847858$ & 9/3 \\
SimNPO / MemFlex & 1.776698 & 1.271154 & $+0.505544$ & 11/1 \\
SimNPO / FILA & 1.776698 & 1.205905 & $+0.570793$ & 8/4 \\
SimNPO / Activation-Down & 1.776698 & 1.273278 & $+0.503420$ & 12/0 \\
\bottomrule
\end{tabular}
\end{adjustbox}

\vspace{0.15\baselineskip}
\setlength{\tabcolsep}{2.2pt}
\begin{adjustbox}{max width=\textwidth}
\begin{tabular}{@{}llcccl@{}}
\toprule
\multicolumn{6}{@{}l}{Panel B: GradDiff three-field sensitivity, Birth City excluded} \\
Baseline & Mean $J$ & DIR-R$-$baseline & Median gap & W/T/L & LOO range \\
\midrule
Projected Causal & 106.577433 & $+0.014444$ & $-0.940000$ & 4/0/5 & $[-1.6425,+1.0056]$ \\
Activation-Down & 104.534099 & $+2.057778$ & $+0.030000$ & 5/0/4 & $[+0.5869,+2.8506]$ \\
Storage & 103.033544 & $+3.558333$ & $+2.150000$ & 7/0/2 & $[+2.2525,+4.7450]$ \\
FILA & 102.956044 & $+3.635833$ & $+1.430000$ & 7/0/2 & $[+1.9897,+4.2047]$ \\
GRAIL & 102.482155 & $+4.109722$ & $+3.150000$ & 8/0/1 & $[+2.3316,+4.6578]$ \\
MemFlex & 102.189655 & $+4.402222$ & $+1.795000$ & 6/0/3 & $[+3.2206,+5.4519]$ \\
WAGLE & 101.389099 & $+5.202778$ & $+4.070000$ & 7/0/2 & $[+4.1113,+6.2088]$ \\
AP-MLP & 99.490766 & $+7.101111$ & $+5.315000$ & 9/0/0 & $[+5.5400,+7.8356]$ \\
\bottomrule
\end{tabular}
\end{adjustbox}
\end{table}

\paragraph{Results.}
\textbf{\textit{Our method has higher mean terminal utility and more run-level wins than losses in all six direct comparisons.}} NPO margins range from $+0.431$ to $+0.848$, and SimNPO margins range from $+0.503$ to $+0.571$.

\paragraph{Three-Field Sensitivity.}
In the three-field sensitivity excluding Birth City, \DIRR{} has the highest mean terminal utility, but its $+0.014444$ margin over Projected Causal is unstable: median $-0.94$, W/T/L $4/0/5$, and LOO $[-1.6425,+1.0056]$.

\paragraph{Birth City.}
The four-field primary remains in Appendix~\ref{app:graddiff}. Projected Causal has a mean terminal utility $7.104583$ higher than \DIRR{}, and all eight external parameter-selection methods have higher Birth City means, with gaps from $-28.461667$ to $-5.872500$ (Appendix Tables~\ref{tab:app-graddiff-primary} and~\ref{tab:app-graddiff-fieldgap}). \textbf{\textit{Most gaps outside Birth City reverse sign, so this field drives the aggregate reversal.}}

\paragraph{Replication.}
A three-field rerun on new seeds reverses the near-tie: Projected Causal leads \DIRR{} by $0.861389$ with the same $4/0/5$ W/T/L (Appendix Table~\ref{tab:app-graddiff-replication}). We therefore treat the external ordering as seed-sensitive rather than as a stable three-field ranking.

\FloatBarrier
\subsection{Ablation: Selective Revision}
\label{sec:gate-results}

\paragraph{Question.}
The DIR-R ablation asks whether the gate can preserve the fixed-support solution while using substantially less step-equivalent compute than the $16\times$ exhaustive multi-checkpoint reference. \StaticIV{} is the $1\times$ fixed-support baseline; \DIRR{} changes only whether support is reconsidered at the checkpoint. Table~\ref{tab:static-dirr-ablation} collects the main Natural-TOFU and LACUNA DIR-R evaluation sets.

\begin{table}[!t]
\centering
\caption{Selective revision relative to \StaticIV{}. Available gain is the improvement from exhaustive candidate evaluation at the frozen decision checkpoint; it is distinct from the $16\times$ exhaustive multi-checkpoint reference.}
\label{tab:static-dirr-ablation}
\footnotesize
\setlength{\tabcolsep}{3.2pt}
\renewcommand{\arraystretch}{0.96}
\begin{adjustbox}{max width=\textwidth}
\begin{tabular}{@{}lllcccc@{}}
\toprule
Benchmark & Objective / setting & $n$ & $\Delta J_{\mathrm{DIRR}}$ & Trigger & Avail. gain & Captured \\
\midrule
Natural-TOFU & NPO & 8 & $+0.002010$ & 50.0\% & $+0.005667$ & 35.5\% \\
Natural-TOFU & SimNPO & 8 & $+0.005130$ & 25.0\% & $+0.013162$ & 39.0\% \\
Natural-TOFU$^{\ddagger}$ & GradDiff & 20 & $+0.004298$ & 40.0\% & $+0.014793$ & 29.1\% \\
\midrule
LACUNA & NPO & 12 & $+0.002433$ & 33.3\% & $+0.008024$ & 30.3\% \\
LACUNA & SimNPO & 12 & $+0.008228$ & 58.3\% & $+0.014868$ & 55.3\% \\
LACUNA & GradDiff, four-field & 12 & $+0.164583$ & 50.0\% & $+0.282500$ & 58.3\% \\
\bottomrule
\end{tabular}
\end{adjustbox}
\vspace{0.05\baselineskip}
\parbox{0.96\linewidth}{\scriptsize $^{\ddagger}$Natural GradDiff uses held-out trace replay. The GradDiff candidate-set denominator was completed post hoc without changing the frozen DIR-R decisions or threshold.}
\end{table}

\paragraph{Safety.}
The $\rho=0$ candidate is exactly the \StaticIV{} continuation. Non-triggered runs retain it, and triggered exhaustive selection includes it among the candidates; when all branches complete, terminal utility cannot decrease relative to \StaticIV{}. \textbf{\textit{The primary four-field GradDiff evaluation has mean gain $+0.164583$, median paired gain $+0.002500$, and six wins, six ties, and no losses.}} The difference between the mean and median shows that gain magnitude is heterogeneous across runs.

\paragraph{Efficiency.}
The main question is how much useful revision is retained for far less computation, not whether every absolute gain is large. Under the GradDiff cost model, non-trigger and trigger paths cost $1.10\times$ and $1.80\times$; with 6/12 triggers, the primary four-field evaluation averages $1.45\times$, below one tenth of the $16\times$ exhaustive multi-checkpoint reference in normalized step-equivalent compute. Across the six evaluation sets with complete candidate-set denominators, \DIRR{} captures from $29.1\%$ to $58.3\%$ of the gain available from exhaustive candidate evaluation at the frozen checkpoint. The primary four-field evaluation captures $58.3\%$ ($+0.164583$ of $+0.282500$). Its denominator was completed only after the \DIRR{} decisions were frozen and is used for diagnosis, not retuning. Appendix~\ref{app:graddiff-raw} retains the three-field replication as a boundary analysis rather than a headline result.

\section{Discussion and Limitations}
\label{sec:discussion}

Our results distinguish parameter localization from the downstream decision of where an unlearning objective should act. The controlled experiment makes this distinction explicit: the injected interface can be identified accurately, yet this information does not predict the better intervention. The parameter-selection comparisons further show that intervention quality depends on the objective and data field. These results do not imply that localization is uninformative; rather, they indicate that localization evidence alone is not a sufficient decision criterion for selecting an editable subset.

The evidence also has several important boundaries. Natural-TOFU margins for Intervention Score + DIR-R combine results from different final evaluation sets and therefore are not direct paired estimates. Natural GradDiff uses held-out trace replay rather than the confirmation evaluation set. For LACUNA GradDiff, the four-field result is the primary endpoint: several static parameter-selection methods remain ahead of DIR-R, and the three-field analysis excluding Birth City is a sensitivity analysis. We therefore interpret it as evidence of field heterogeneity rather than as a replacement superiority result.

Finally, \DIRR{} searches only within a frozen family of equal-budget continuation candidates. Its exhaustive comparisons identify the best continuation within that family, not the globally optimal editable subset. The gate trades additional computation for continuation gain: the $16\times$ exhaustive multi-checkpoint construction is a normalized step-equivalent reference rather than a wall-clock measurement, and the appropriate balance between compute and gain may vary across objectives and deployment settings.

\section{Conclusion}
\label{sec:conclusion}

Localized LLM unlearning involves two decisions that are often conflated: which parameters should be updated initially, and whether that choice should remain fixed as optimization changes the model. We formalized these decisions separately. Intervention Score supplies an objective-conditioned initial support, while selective \DIRR{} tests whether a checkpoint justifies revising that support.

The evaluations clarify both the value and the limits of this decomposition. The controlled study shows that accurate storage localization can fail to identify the better intervention. Natural-TOFU provides broad descriptive evidence across ten parameter-selection baselines, while the direct LACUNA comparisons show positive mean margins under NPO and SimNPO. The GradDiff results add an important qualification: external static rankings vary by data field and seed, and several static methods remain ahead in the four-field primary. Against its own \StaticIV{} reference, however, \DIRR{} has a positive aggregate mean difference in every reported evaluation set while invoking the complete continuation comparison only after a trigger. Localized unlearning should therefore treat support selection as a decision about the effect of an update, report where that decision is unstable, and evaluate checkpoint-dependent revision separately from initial localization quality. This framing permits stronger claims where comparisons are direct without hiding boundary cases where method orderings change. Future work should test learned candidate families and multi-checkpoint gates under stronger forgetting attacks and measured wall-clock constraints. The central question is not only where target information is found, but whether updating those parameters improves forgetting without unacceptable collateral damage at the current model state. This makes support control a distinct design and evaluation problem.

\clearpage
\section*{AI Use Statement}

Generative AI tools were used for literature search and related-work discovery, feedback on experimental design and methodology, experimental implementation including code generation, analysis and interpretation of experimental results, and drafting and polishing parts of the manuscript. All AI-assisted content, code, references, and experimental results were reviewed and verified by the authors, who take responsibility for the final paper.

\section*{Reproducibility Statement}

Section~\ref{sec:method} and Algorithm~\ref{alg:dirr} specify Intervention Score, the fixed-subset \StaticIV{} reference, and the calibration and deployment procedures for selective \DIRR{}. Appendix~\ref{app:method-details} documents support-constrained optimization, checkpoint restoration, candidate construction, and threshold calibration. Appendix~\ref{app:experimental-details} records the comparison protocols, evaluation metrics, aggregation and missingness rules, and the controlled separation between storage and intervention. Appendix~\ref{app:complete-results} reports the complete audited result tables, including the GradDiff \DIRR{} analysis, candidate-set coverage, and replication analyses. An anonymized supplementary code package is organized by the table it supports.

\clearpage

\bibliographystyle{iclr2027_conference}
\bibliography{references}

\clearpage
\appendix

\section{Methodological Details}
\label{app:method-details}

\subsection{Core Notation}

\paragraph{Notation.}
Table~\ref{tab:notation} summarizes symbols already defined in Sections~\ref{sec:problem} and~\ref{sec:method}.

\begin{table}[H]
\centering
\caption{Core notation used in the problem formulation and method.}
\label{tab:notation}
\papertableformat
\begin{tabular}{p{0.12\textwidth}p{0.34\textwidth}p{0.12\textwidth}p{0.34\textwidth}}
\toprule
Symbol & Meaning & Symbol & Meaning \\
\midrule
$f_{\vtheta}$, $\vtheta_0$ & Model and pre-unlearning parameters. & $T$, $t$ & Total steps and DIR-R decision checkpoint. \\
$\mathcal D_f,\mathcal D_r$ & Forget and retain sets. & $\mathcal L_{\mathrm{unl}}$ & Fixed unlearning objective. \\
$\Gset$, $g$ & Editable groups and one group. & $\vtheta_g$ & Parameter coordinates in group $g$. \\
$\Sset$ & Active editable support. & $c(\Sset)$, $B$ & Editable-parameter cost and budget. \\
$\mathcal F_B$ & Supports satisfying $c(\Sset)\le B$. & $J(\vtheta)$ & Forgetting and retention evaluation utility. \\
$V_0(\Sset)$ & Value of using $\Sset$ from the start. & $\Sset_0^\star$, $\Sset_0$ & Ideal and actually selected initial supports. \\
$\mathcal C_t$ & Restorable training checkpoint saved at step $t$. & $\vtheta_T(\Sset;\mathcal C_t)$ & Terminal parameters after continuing from $\mathcal C_t$ with support $\Sset$. \\
$\vtheta_{t+d}(\Sset;\mathcal C_t)$ & Parameters after $d$ continuation steps from $\mathcal C_t$. & $V_t(\Sset\mid\mathcal C_t)$ & Terminal continuation value from $\mathcal C_t$. \\
$\Sset_t^\star(\mathcal C_t)$ & Ideal support after observing $\mathcal C_t$. & $\mP_{\Sset}$ & Projector onto support coordinates. \\
$\mathbf u_k^{\Sset}$, $\eta_k$ & Support-conditioned optimizer direction and step size. & $e_{g,q}$ & First-order diagnostic effect, $q\in\{F,R,P,N\}$. \\
$\rho$, $\Aset$ & Replacement fraction and frozen action set. & $\rho_p$, $d$ & Probe action and probe length. \\
$\widetilde V_t^{(d)}(\rho\mid\mathcal C_t)$ & Short-horizon probe score. & $s_t$, $\tau^\star$ & Gate signal and frozen threshold. \\
\bottomrule
\end{tabular}
\end{table}

\subsection{Intervention Score Details}
\label{app:intervention-score-details}
Intervention Score uses gradients only at the initial model. The objective direction $\mathbf u_0$ is computed before optimizer preconditioning so that the score reflects the declared objective rather than optimizer-specific moment estimates. Each diagnostic gradient $\mathbf g_q$ is evaluated on its own frozen diagnostic records at the same $\vtheta_0$. No terminal model score, test outcome, or later checkpoint is used to construct $s_{\mathrm{int}}$.

Equation~\ref{eq:intervention-taylor} gives the local interpretation of $e_{g,q}$. The four roles separate the intended effect from two forms of collateral damage and from nonspecific movement: $F$ measures the forget-directed effect, $R$ ordinary retention, $P$ protected behavior outside the ordinary retain diagnostic, and $N$ a neutral control. The score in Eq.~\ref{eq:intervention-score} is used only for ordering groups. Once $\Sset_0$ is selected, the actual training update is the original unlearning objective restricted to the active coordinates; Intervention Score is not added to the training loss.

For the equal-cost column geometry used in the primary experiments, selecting $\Sset_0$ amounts to sorting groups by $s_{\mathrm{int}}$ and taking the largest prefix whose editable-parameter count does not exceed $B$. If group costs differ, the same definition of $c(\Sset)$ applies, but the packing routine must enforce the budget explicitly rather than assuming a fixed number of groups.

\subsection{Support-Constrained Optimization}

\paragraph{Updates.}
For support $\Sset$, $\mP_{\Sset}$ projects onto the coordinates in its groups. If the fixed objective and optimizer produce direction $\mathbf u_k^{\Sset}$ and step size $\eta_k$ at step $k$, support-constrained optimization applies
\begin{equation}
\vtheta_{k+1}^{\Sset}
=\vtheta_k^{\Sset}-\eta_k\mP_{\Sset}\mathbf u_k^{\Sset}.
\label{eq:appendix-support-update}
\end{equation}
All coordinates outside $\Sset$ are held fixed. Restoring checkpoint $\mathcal C_t$ also reinstates the optimizer and scheduler state, the remaining minibatch position, and the random-number-generator state, so continuation candidates differ only in editable support.

\subsection{DIR-R Candidate Construction And Calibration}

\paragraph{Candidates.}
Let $M=|\Sset_0|$. DIR-R orders selected groups from lowest to highest Intervention Score and unselected groups in the reverse order. For $k(\rho)=\lfloor\rho M\rfloor$, it replaces the first $k(\rho)$ selected groups by the first $k(\rho)$ replacement groups. In the primary equal-cost column geometry this preserves the editable-parameter count exactly. The frozen family is $\Aset=\{0,0.01,0.025,0.05,0.10\}$ and $\rho=0$ is the unchanged \StaticIV{} continuation.

\paragraph{Compute.}
For the $T=200$ exhaustive multi-checkpoint reference, all five candidates are continued from $t\in\{0,40,80,120,160\}$. Counting all partial rollouts gives $5(200+160+120+80+40)=3000$ branch steps; with the committed $200$-step trajectory, normalized compute is $16\times$. This is algorithmic step-equivalent accounting, not measured wall-clock time.

\paragraph{Calibration.}
For development checkpoint $\mathcal C_{i,t}$, define the best candidate continuation score and nonnegative available gain
\begin{align}
V_{i,t}^{\mathrm{best}}
&=\max_{\rho\in\Aset}V_{i,t}(\Sset^{(\rho)}\mid\mathcal C_{i,t}),\\
\gamma_i
&=\max\!\left\{0,
V_{i,t}^{\mathrm{best}}-V_{i,t}(\Sset^{(0)}\mid\mathcal C_{i,t})
\right\}.
\label{eq:dev-gain-app}
\end{align}
For threshold $\tau$, let $\mathcal I_\tau=\{i:s_i>\tau\}$. Among thresholds whose trigger fraction lies in a prespecified admissible interval, $\tau^\star$ maximizes $\sum_{i\in\mathcal I_\tau}\gamma_i$ divided by the corresponding additional step-equivalent continuation cost. Natural-TOFU uses an admissible trigger range from $20\%$ to $80\%$. Final outcomes never alter this threshold or the action family.

The complete procedure now appears in Algorithm~\ref{alg:dirr} in the main paper. Its probe branches and triggered continuation candidates are restored independently from the same checkpoint, preventing optimizer or random-number-generator state from leaking across candidates.

\section{Experimental Details}
\label{app:experimental-details}

\subsection{Comparison Protocols}

\paragraph{Natural-TOFU.}
The Natural-TOFU dataset uses Llama 3 8B and Gemma 2 2B under NPO and SimNPO for the method comparison. All parameter-selection methods act in the same MLP down-projection column space and use approximately six million trainable scalar parameters in the primary setting. Because the parameter-selection panel and \DIRR{} evaluation set are distinct, the reported Natural-TOFU margins combine separately estimated components and are not interpreted as direct paired estimates.

\paragraph{LACUNA.}
LACUNA uses the frozen OLMo 3 7B artifact, four personally identifiable information (PII) fields, and NPO, SimNPO, and GradDiff. NPO and SimNPO baseline identities are fixed before the outcomes are opened and are evaluated on the same field, objective, seed, parameter budget, and evaluator as \DIRR{}. The GradDiff closure uses $-g_{\mathrm{forget}}+g_{\mathrm{retain}}$, $T=200$, a $456{,}862{,}557$-parameter budget, final seeds $4099/8191/16381$, calibration seeds $73/311/997$, decision checkpoint $160$, and a $20$-step probe. \StaticIV{}, \DIRR{}, and the external parameter-selection methods are compared from completed outcomes without retraining or cross-run substitution.

\subsection{Metric Definitions}
\label{app:metric-definitions}

\paragraph{Terminal Score.}
For evaluation run $i$, the evaluator returns forgetting gain $G_{F,i}$ and benchmark-specific nonnegative collateral damage $D_{\mathrm{coll},i}$, with
\begin{equation}
J_i=G_{F,i}-D_{\mathrm{coll},i}.
\label{eq:terminal-score-metric}
\end{equation}
Larger $J$ indicates a better balance between forgetting and utility. For methods $A$ and $B$, $\Delta J(A,B)=n^{-1}\sum_i[J_i(A)-J_i(B)]$; positive values favor $A$. Median gap and W/T/L use the same paired differences. The LOO range is the minimum and maximum mean gap after deleting one run at a time; a range entirely above zero is less dependent on any single run.

\paragraph{DIR-R Metrics.}
Trigger rate is $n^{-1}\sum_i\mathbf 1[s_i>\tau^\star]$ and measures how often exhaustive continuation evaluation is invoked; lower is cheaper but is not intrinsically better. If every final run has outcomes for all $\rho\in\Aset$, the available candidate-set gain is
\begin{equation}
G_{\mathrm{avail}}=\frac{1}{n}\sum_i\left[\max_{\rho\in\Aset}V_{i,t}(\Sset^{(\rho)}\mid\mathcal C_{i,t})-V_{i,t}(\Sset^{(0)}\mid\mathcal C_{i,t})\right].
\label{eq:available-gain-metric}
\end{equation}
The fraction of available gain captured is
\begin{equation}
C_{\mathrm{gain}}=
\frac{\sum_i\left[J_i(\mathrm{DIRR})-J_i(\mathrm{StaticIV})\right]}
{\sum_i\left[\max_{\rho\in\Aset}V_{i,t}(\Sset^{(\rho)}\mid\mathcal C_{i,t})-V_{i,t}(\Sset^{(0)}\mid\mathcal C_{i,t})\right]}.
\label{eq:gain-capture-metric}
\end{equation}
Larger $C_{\mathrm{gain}}$ means that selective revision captures more opportunity within the frozen candidate family. It is undefined when non-trigger final runs lack exhaustive candidate outcomes.

\subsection{Aggregation And Missingness}

\paragraph{Missingness.}
Operationally missing runs are excluded rather than replaced. The only missing run affecting the Natural-TOFU static confirmation is the Gemma 2 2B, NPO, Storage configuration, for which the relevant aggregation uses $n=19$ units. Engineering failures are not treated as negative scientific outcomes. Baseline identities, candidate fractions, calibration and final assignments, and gate thresholds are frozen before final outcomes are inspected.

\subsection{Controlled Storage and Intervention Construction}

\paragraph{Independence.}
The controlled experiment uses synthetic facts injected through two parameter interfaces in Qwen2.5-7B. Localization is evaluated without access to intervention outcomes. Intervention quality is subsequently measured with NPO runs that share the objective, batches, optimization steps, seeds, and approximately $6$M editable scalar parameters. This protocol prevents direct use of downstream intervention outcomes when constructing the localization score.

\section{Complete Audited Results}
\label{app:complete-results}

\subsection{Natural-TOFU Complete Composed Comparison}

\paragraph{Decomposition.}
For baseline $C$, the Natural-TOFU margin for Intervention Score + DIR-R is
\begin{equation}
\Delta J_{\mathrm{full},C}
=
\Delta J_{\mathrm{Static},C}
+
\Delta J_{\mathrm{DIRR},\mathrm{Static}}.
\label{eq:natural-combination-app}
\end{equation}
Because the two terms come from different final evaluation sets, this equation combines separately estimated components rather than defining a paired estimator. Table~\ref{tab:app-natural-complete} exposes both terms instead of reporting only the final margin.

\begin{table}[H]
\centering
\caption{Natural-TOFU audited calculation of the Intervention Score + DIR-R margin. Positive values favor our method. Totals are computed from unrounded values; displayed components are rounded.}
\label{tab:app-natural-complete}
\papertableformat
\begin{adjustbox}{max width=\textwidth}
\begin{tabular}{llccc}
\toprule
Objective & Baseline & $\StaticIV-C$ & DIR-R$-\StaticIV$ & Full method$-C$ \\
\midrule
NPO & Random & +0.002789 & +0.002010 & +0.004799 \\
NPO & Activation & +0.038996 & +0.002010 & +0.041006 \\
NPO & WAGLE & +0.134441 & +0.002010 & +0.136450 \\
NPO & Storage & +0.030793 & +0.002010 & +0.032802 \\
NPO & MemFlex & +0.092770 & +0.002010 & +0.094780 \\
NPO & FILA-relative Fisher & +0.001380 & +0.002010 & +0.003390 \\
NPO & AP-MLP & +0.027762 & +0.002010 & +0.029772 \\
NPO & GRAIL & +0.000163 & +0.002010 & +0.002172 \\
NPO & Activation-Down & +0.037585 & +0.002010 & +0.039595 \\
NPO & Projected Causal & +0.108938 & +0.002010 & +0.110948 \\
\midrule
SimNPO & Random & +0.001844 & +0.005130 & +0.006974 \\
SimNPO & Activation & +0.149612 & +0.005130 & +0.154743 \\
SimNPO & WAGLE & +0.292638 & +0.005130 & +0.297768 \\
SimNPO & Storage & +0.157138 & +0.005130 & +0.162268 \\
SimNPO & MemFlex & +0.205061 & +0.005130 & +0.210191 \\
SimNPO & FILA-relative Fisher & -0.005930 & +0.005130 & -0.000799 \\
SimNPO & AP-MLP & +0.093010 & +0.005130 & +0.098141 \\
SimNPO & GRAIL & -0.004412 & +0.005130 & +0.000719 \\
SimNPO & Activation-Down & +0.144147 & +0.005130 & +0.149278 \\
SimNPO & Projected Causal & +0.126655 & +0.005130 & +0.131786 \\
\bottomrule
\end{tabular}
\end{adjustbox}
\end{table}

\paragraph{Boundary Cases.}
Under NPO, \StaticIV{} already has a positive margin over every baseline, so the $+0.002010$ DIR-R increment increases rather than reverses the ordering. SimNPO is more informative. \StaticIV{} trails both FILA-relative Fisher by $0.005930$ and GRAIL by $0.004412$. Adding the $+0.005130$ DIR-R increment is sufficient to move GRAIL to a small positive margin for Intervention Score + DIR-R of $+0.000719$, but not sufficient to overturn FILA-relative Fisher, which remains ahead by $0.000799$. The dynamic component therefore contributes a positive descriptive increment to the reported margin while leaving a visible boundary on the method-level claim.

\subsection{LACUNA NPO/SimNPO Parameter-Selection Results}

\paragraph{Objective Sensitivity.}
Table~\ref{tab:app-lacuna-selection-comparison} reports the ten-method parameter-selection comparison. Under NPO, Projected Causal has the highest mean $J$ at $6.538359$, followed by MemFlex at $6.374923$ and WAGLE at $6.253223$. Under SimNPO, MemFlex becomes the best-performing parameter-selection method at $1.745046$, FILA rises to $1.605025$, while Projected Causal falls to $1.331232$. In rank terms, Projected Causal moves from first under NPO to eighth under SimNPO, whereas FILA moves from eighth to second. The same editable geometry therefore produces materially different parameter-selection rankings when only the unlearning objective changes.

\begin{table}[H]
\centering
\caption{LACUNA NPO/SimNPO parameter-selection comparison. Each mean for a method and objective uses $n=12$ unique runs after deduplication.}
\label{tab:app-lacuna-selection-comparison}
\papertableformat
\begin{tabular}{lcc}
\toprule
Baseline & NPO mean $J$ & SimNPO mean $J$ \\
\midrule
Random & 5.105164 & 1.035321 \\
Activation exact & 6.174900 & 1.586493 \\
WAGLE & 6.253223 & 1.470818 \\
Storage & 5.839900 & 1.489109 \\
MemFlex & 6.374923 & 1.745046 \\
FILA & 5.362761 & 1.605025 \\
AP-MLP & 5.088650 & 1.316195 \\
GRAIL & 5.614737 & 1.580461 \\
Activation-Down & 6.169626 & 1.585372 \\
Projected Causal & 6.538359 & 1.331232 \\
\bottomrule
\end{tabular}
\end{table}

\paragraph{Static Ordering.}
The objective-dependent rank reversals are larger than a simple exchange between two near-tied methods. Projected Causal is the best-performing NPO parameter-selection method but is below seven alternatives under SimNPO, while FILA exhibits the opposite pattern. MemFlex is comparatively stable, ranking second under NPO and first under SimNPO. These differences support conditioning initial parameter selection on the actual unlearning objective rather than transferring one static ranking across objectives.

\subsection{LACUNA GradDiff Four-Field Primary And Field Heterogeneity}
\label{app:graddiff}

\paragraph{Four-Field Primary.}
The GradDiff closure uses $-g_{\mathrm{forget}}+g_{\mathrm{retain}}$ and contains 12 outcomes covering Birth City, Driver's License, Email Address, and Phone Number with three final seeds per field. In the four-field aggregate, \DIRR{} has mean $J=102.628354$. Projected Causal, Activation-Down, FILA, GRAIL, and MemFlex have higher means, while Storage, WAGLE, and AP-MLP have lower means. The largest negative mean gap is $-7.104583$ against Projected Causal, whereas the largest positive gap is $+3.857708$ against AP-MLP. The primary result therefore establishes neither universal superiority nor universal inferiority.

\begin{table}[H]
\centering
\caption{GradDiff four-field primary result. Positive DIR-R gaps favor DIR-R.}
\label{tab:app-graddiff-primary}
\papertableformat
\begin{adjustbox}{max width=\textwidth}
\begin{tabular}{lllll}
\toprule
Method & Mean $J$ & DIR-R$-$method & Median paired gap & DIR-R W/T/L \\
\midrule
Projected Causal & 109.732938 & -7.104583 & -1.412500 & 4/0/8 \\
Activation-Down & 105.319396 & -2.691042 & -0.030000 & 6/0/6 \\
FILA & 104.562938 & -1.934583 & +0.620000 & 7/0/5 \\
GRAIL & 103.815438 & -1.187083 & +1.072500 & 8/0/4 \\
MemFlex & 103.104396 & -0.476042 & +1.105000 & 7/0/5 \\
DIR-R & 102.628354 & N/A & N/A & N/A \\
\StaticIV{} & 102.463771 & +0.164583 & N/R & 6/6/0 \\
Storage & 102.227313 & +0.401042 & +2.020000 & 8/0/4 \\
WAGLE & 100.976063 & +1.652292 & +2.880000 & 8/0/4 \\
AP-MLP & 98.770646 & +3.857708 & +3.997500 & 10/0/2 \\
\bottomrule
\end{tabular}
\end{adjustbox}
\end{table}

\paragraph{Birth City.}
The field decomposition in Table~\ref{tab:app-graddiff-fieldgap} shows a common pattern across all eight external parameter-selection methods: every Birth City gap is negative, ranging from $-5.872500$ against AP-MLP to $-28.461667$ against Projected Causal. Outside Birth City, most gaps reverse sign. For example, DIR-R trails FILA by $18.645833$ on Birth City but leads it by $5.211667$, $3.740000$, and $1.955833$ on Driver's License, Email Address, and Phone Number. Similar sign reversals occur for GRAIL, Storage, WAGLE, and AP-MLP. The four-field aggregate therefore hides substantial variation across data fields.

\begin{table}[H]
\centering
\caption{GradDiff field-level DIR-R gaps. Positive values favor DIR-R.}
\label{tab:app-graddiff-fieldgap}
\papertableformat
\begin{adjustbox}{max width=\textwidth}
\begin{tabular}{lllll}
\toprule
Baseline & Birth City & Driver's License & Email Address & Phone Number \\
\midrule
Projected Causal & -28.461667 & -0.651667 & +0.855000 & -0.160000 \\
Activation-Down & -16.937500 & +1.838333 & +4.823333 & -0.488333 \\
FILA & -18.645833 & +5.211667 & +3.740000 & +1.955833 \\
GRAIL & -17.077500 & +7.231667 & +3.265000 & +1.832500 \\
MemFlex & -15.110833 & +7.693333 & +5.926667 & -0.413333 \\
Storage & -9.070833 & +6.756667 & +3.368333 & +0.550000 \\
WAGLE & -8.999167 & +8.260000 & +6.793333 & +0.555000 \\
AP-MLP & -5.872500 & +9.175000 & +10.338333 & +1.790000 \\
\bottomrule
\end{tabular}
\end{adjustbox}
\end{table}

\paragraph{Three-Field Sensitivity.}
Section~\ref{sec:lacuna-main} reports the three-field sensitivity obtained by removing the already-completed Birth City runs. That sensitivity changes the aggregate ordering but does not replace the four-field primary above. The replication below further tests whether the apparent three-field near-tie with Projected Causal persists.

\subsection{GradDiff Raw DIR-R Audit And Replication}
\label{app:graddiff-raw}

\paragraph{Field Decomposition.}
Table~\ref{tab:app-graddiff-dirr-field} decomposes the primary four-field GradDiff \DIRR{} result by field. \DIRR{} has a higher mean $J$ than \StaticIV{} in each field, including Birth City. The large Birth City discrepancy in the external parameter-selection comparison therefore does not arise because \DIRR{} underperforms its own \StaticIV{} baseline on that field; it arises because several alternative static parameter-selection methods attain substantially higher Birth City scores.

\begin{table}[H]
\centering
\caption{GradDiff primary four-field \DIRR{} decomposition. The final column gives triggered runs among the three final seeds.}
\label{tab:app-graddiff-dirr-field}
\papertableformat
\begin{tabular}{lllcl}
\toprule
Field & \StaticIV{} $J$ & DIR-R $J$ & DIR-R$-\StaticIV$ & Triggered \\
\midrule
Birth City & 90.536120 & 90.737786 & $+0.201667$ & 1/3 \\
Driver's License & 111.466042 & 111.839375 & $+0.373333$ & 2/3 \\
Email Address & 118.499479 & 118.559479 & $+0.060000$ & 1/3 \\
Phone Number & 89.353444 & 89.376777 & $+0.023333$ & 2/3 \\
\midrule
Overall & 102.463771 & 102.628354 & $+0.164583$ & 6/12 \\
\bottomrule
\end{tabular}
\end{table}

\paragraph{Gate Behavior.}
The primary GradDiff threshold is frozen at $\tau=0.02125$ from 12 calibration instances spanning four fields and seeds $73/311/997$. Table~\ref{tab:app-graddiff-calibration} summarizes the calibration outcome. All five triggered calibration decisions have positive available gain, but the gate does not identify every beneficial revision: for example, Birth City seed 311 and Driver's License seed 997 have zero probe signal while their best candidate-set gains are $+0.52$ and $+0.22$, respectively. The gate is therefore a conservative screening rule rather than a predictor of every continuation opportunity.

\begin{table}[H]
\centering
\caption{GradDiff primary calibration summary. Gain captured is the fraction of total positive available gain over the frozen continuation candidate set.}
\label{tab:app-graddiff-calibration}
\papertableformat
\begin{tabular}{ll}
\toprule
Quantity & Value \\
\midrule
Frozen threshold $\tau$ & 0.021250 \\
Calibration instances & 12 \\
Triggered instances & 5/12 \\
Trigger rate & 41.67\% \\
Trigger precision & 1.000 \\
Captured positive gain & 0.9025 \\
Fraction of positive available gain captured & 0.476882 \\
Captured gain / relative extra compute & 0.192021 \\
\bottomrule
\end{tabular}
\end{table}

\paragraph{Candidate-Set Coverage.}
The primary four-field evaluation and the three-field replication stored exhaustive five-candidate outcomes only for triggered runs (6/12 and 2/9). A post-hoc supplement evaluated the missing non-trigger branches without changing the frozen threshold, support choices, or original \DIRR{} actions, yielding complete 12/12 and 9/9 candidate-set denominators. The primary evaluation has mean available candidate-set gain $+0.282500$ and captures $58.26\%$ of it; the replication has mean available gain $+0.183333$ and captures $5.45\%$. These are single-checkpoint candidate-set quantities, not quantities normalized to the $16\times$ exhaustive multi-checkpoint reference.

\paragraph{Replication.}
The three-field replication reruns Driver's License, Email Address, and Phone Number on seeds $32771/65537/131071$, keeps the primary threshold $\tau=0.02125$, and performs no recalibration. All nine \DIRR{}/\StaticIV{} runs and all nine Projected Causal runs complete.

\begin{table}[H]
\centering
\caption{GradDiff three-field sensitivity and replication. The first row is derived from the completed primary runs; the second uses new seeds.}
\label{tab:app-graddiff-replication}
\papertableformat
\begin{adjustbox}{max width=\textwidth}
\begin{tabular}{llcccccccc}
\toprule
Setting & Seeds & \StaticIV{} & DIR-R & DIR-R$-$Static & Projected & DIR-R$-$Projected & Median gap & W/T/L & Trigger \\
\midrule
Original three-field & 4099/8191/16381 & 106.439655 & 106.591877 & $+0.152222$ & 106.577433 & $+0.014444$ & $-0.940000$ & 4/0/5 & 5/9 \\
Replication & 32771/65537/131071 & 107.729099 & 107.739099 & $+0.010000$ & 108.600488 & $-0.861389$ & $-1.440000$ & 4/0/5 & 2/9 \\
\bottomrule
\end{tabular}
\end{adjustbox}
\end{table}

\paragraph{Replication Outcome.}
The original three-field sensitivity places DIR-R only $+0.014444$ above Projected Causal in mean score, with a negative paired median and 4/0/5 W/T/L. In the replication, Projected Causal instead has the higher mean by $0.861389$, while W/T/L remains 4/0/5. The paired gaps are highly variable across seeds; exposed examples include $+16.975$ and $-10.310$ on different Driver's License seeds, $-6.725$ and $-7.205$ on two Email Address seeds, and $+2.8425$ on one Phone Number seed. Thus the external ordering is not stable enough to support a three-field superiority claim.

\paragraph{DIR-R.}
The primary four-field evaluation has mean paired gain $+0.164583$, median $+0.002500$, and 6/6/0 W/T/L across its 12 final runs. The completed denominator shows that this corresponds to $58.26\%$ of the $+0.282500$ mean gain available from exhaustive candidate evaluation at the frozen checkpoint. This quantifies the fraction of candidate-set revision retained by \DIRR{}. In the three-field replication, only Email Address seed 32771 and Phone Number seed 32771 trigger, with gains of $+0.05$ and $+0.04$; the remaining seven runs return the \StaticIV{} continuation. The aggregate gain is $+0.010000$ against $+0.183333$ available candidate-set gain, or $5.45\%$, so we retain the replication as a near-tie boundary result rather than evidence of a substantial \DIRR{} gain.

\subsection{LACUNA Fixed-Subset Analyses}

\paragraph{Static-IV.}
On the canonical NPO/SimNPO parent panel, \StaticIV{} has combined mean $J=4.187641$. The corresponding means are $3.070243$ for Random, $3.862021$ for WAGLE, $3.664504$ for Storage, and $4.059985$ for MemFlex, producing paired gaps of $+1.117398$, $+0.325620$, $+0.523136$, and $+0.127656$. The MemFlex descriptive interval crosses zero, so the smallest of these four gaps should not be interpreted as a robust separation.

\begin{table}[H]
\centering
\caption{LACUNA analysis for modern parameter-selection methods.}
\label{tab:app-lacuna-modern}
\papertableformat
\begin{adjustbox}{max width=\textwidth}
\begin{tabular}{lcccl}
\toprule
Baseline & Baseline mean $J$ & $\StaticIV-C$ & Run wins & Setting wins / 95\% CI \\
\midrule
FILA-relative Fisher & 3.4839 & +0.7037 & 36/48 & 14/16 $[+0.3765,+1.0244]$ \\
AP-MLP & 3.2024 & +0.9852 & 44/48 & 16/16 $[+0.7097,+1.2726]$ \\
GRAIL & 3.5976 & +0.5900 & 39/48 & 14/16 $[+0.3185,+0.8568]$ \\
Activation-Down & 3.8775 & +0.3101 & 37/48 & 13/16 $[+0.0491,+0.5454]$ \\
Projected Causal & 3.9348 & +0.2528 & 35/48 & 11/16 $[-0.1109,+0.6112]$ \\
\bottomrule
\end{tabular}
\end{adjustbox}
\end{table}

\paragraph{Modern Parameter-Selection Methods.}
\StaticIV{} has positive mean gaps against all five modern parameter-selection methods in Table~\ref{tab:app-lacuna-modern}. The largest is $+0.9852$ against AP-MLP, accompanied by 44/48 run wins and 16/16 setting wins. The intervals against FILA-relative Fisher, AP-MLP, GRAIL, and Activation-Down remain positive. Projected Causal is closer: the mean gap is $+0.2528$, but its $95\%$ interval $[-0.1109,+0.6112]$ crosses zero. The static evidence therefore supports using Intervention Score as a competitive fixed-subset baseline without establishing separation from every parameter-selection method near the frontier.

\subsection{LACUNA DIR-R Comparison}

\paragraph{Comparison.}
Table~\ref{tab:app-lacuna-comparison} contains the six comparisons used for the main direct method comparison. Every mean gap is positive. NPO improvements range from $+0.430752$ to $+0.847858$, while SimNPO improvements range from $+0.503420$ to $+0.570793$. Every baseline also loses a majority of its 12 runs, with the strongest run-level result occurring against Activation-Down under SimNPO at 12/0.

\begin{table}[H]
\centering
\caption{LACUNA comparison of Intervention Score + DIR-R with the selected baselines.}
\label{tab:app-lacuna-comparison}
\papertableformat
\begin{tabular}{llcccl}
\toprule
Objective & Baseline & Baseline $J$ & DIR-R $J$ & DIR-R$-C$ & Wins/Losses \\
\midrule
NPO & Projected Causal & 5.763156 & 6.193908 & +0.430752 & 8/4 \\
NPO & MemFlex & 5.502749 & 6.193908 & +0.691159 & 10/2 \\
NPO & Activation exact & 5.346050 & 6.193908 & +0.847858 & 9/3 \\
SimNPO & MemFlex & 1.271154 & 1.776698 & +0.505544 & 11/1 \\
SimNPO & FILA-relative Fisher & 1.205905 & 1.776698 & +0.570793 & 8/4 \\
SimNPO & Activation-Down & 1.273278 & 1.776698 & +0.503420 & 12/0 \\
\bottomrule
\end{tabular}
\end{table}

\paragraph{Evidence Strength.}
The LACUNA results provide a direct comparison between Intervention Score + DIR-R and the selected baselines because both sides are evaluated within the same evaluation settings. Natural-TOFU has a different evidential status because its static and dynamic components come from different final evaluation sets.

\subsection{LACUNA DIR-R Ablation}

\paragraph{Efficiency.}
Under NPO, \DIRR{} increases $J$ from $1.325934$ to $1.328368$, a gain of $+0.002433$, while triggering on $33.3\%$ of runs and producing a wall-time ratio of $1.33\times$. SimNPO increases from $1.680125$ to $1.688353$, a gain of $+0.008228$, with a $58.3\%$ trigger rate and a $1.52\times$ wall-time ratio. These rows capture $30.32\%$ and $55.34\%$ of the corresponding available candidate-set gain. The post-hoc GradDiff supplement completes the same denominator for the primary evaluation and replication without altering their frozen \DIRR{} decisions.

\begin{table}[H]
\centering
\caption{LACUNA DIR-R versus the fixed-subset \StaticIV{} continuation. Available gain and capture use exhaustive candidates at the frozen decision checkpoint.}
\label{tab:app-lacuna-dirr}
\papertableformat
\begin{adjustbox}{max width=\textwidth}
\begin{tabular}{lllcccl}
\toprule
Objective / setting & \StaticIV{} $J$ & DIR-R $J$ & Difference & Trigger & Avail. gain & Captured \\
\midrule
NPO & 1.325934 & 1.328368 & +0.002433 & 33.3\% & +0.008024 & 30.32\% \\
SimNPO & 1.680125 & 1.688353 & +0.008228 & 58.3\% & +0.014868 & 55.34\% \\
GradDiff, four-field & 102.463771 & 102.628354 & +0.164583 & 50.0\% & +0.282500 & 58.26\% \\
GradDiff, three-field replication & 107.729099 & 107.739099 & +0.010000 & 22.2\% & +0.183333 & 5.45\% \\
\bottomrule
\end{tabular}
\end{adjustbox}
\end{table}

\paragraph{GradDiff Replication.}
The primary four-field evaluation triggers on $50.0\%$ of its 12 runs. All six triggered runs improve, giving six wins, six ties, and no losses over all 12 runs; the mean triggered gain is $+0.329167$ and the median is $+0.365$. After the frozen decisions, the missing non-trigger continuations were evaluated only to complete the denominator: the primary evaluation captures $58.26\%$ of the available candidate-set gain. The three-field replication triggers on 2/9 runs and captures only $5.45\%$, consistent with a more conservative gate in that evaluation.

\paragraph{Audit Summary.}
The audited results support three distinct empirical conclusions. First, Intervention Score has positive mean gaps against several parameter-selection methods, but neither Natural-TOFU nor GradDiff supports universal static dominance. Second, LACUNA NPO/SimNPO has positive mean margins in all six reported baseline comparisons. Third, the aggregate DIR-R-minus-\StaticIV{} mean is positive in each reported DIR-R evaluation set, including both GradDiff evaluation sets in which the external static parameter-selection ordering is field- and seed-sensitive. The data therefore support a decomposition between initial intervention selection and checkpoint-dependent revision rather than a claim that one parameter-selection method or DIR-R dominates every setting.

\end{document}